\documentclass[unnumsec,webpdf,traditional,large]{oup-authoring-template-adjusted}
\onecolumn % for one column layouts

\graphicspath{{Fig/}}

\theoremstyle{thmstyleone}%
\theoremstyle{thmstyletwo}%
\theoremstyle{thmstylethree}%

\newcommand{\eg}{\emph{e.g.},\ }

\usepackage{booktabs}
\usepackage{subfig}
\usepackage{hyperref}
\usepackage[printonlyused]{acronym}
\usepackage{setspace}
\usepackage{booktabs}
\usepackage{multirow}
\usepackage{pdfpages}

\acrodef{ICD}{{International Classification of Diseases}}
\acrodef{NLP}{{Natural Language Processing}}
\acrodef{LMTC}{{Large-Scale Multi-Labelled Text Classification}}
\acrodef{NER}{{Named Entity Recognition}}
\acrodef{WHCM}{{Weak Hierarchical Confusion Matrix}}
\acrodef{CoPHE}{{Count-Preserving Hierarchical Evaluation}}
\acrodef{LLMs}{{Large Language Models}}
\title{Can GPT-3.5 Generate and Code Discharge Summaries?}

\newcommand{\mf}[1]{{#1}}

\begin{document}
% Word count: up to 4000 words.
% Structured abstract: up to 250 words.
% Tables: up to 4.
% Figures: up to 6.
% References: unlimited.

% Accepted manuscripts will be required to have a patient/community facing abstract that highlights key findings.

% The manuscript should be presented in the following order:

%     Title page.
%     Abstract, or a summary for case reports (Note: references should not be included in abstracts or summaries).
%     Main text separated under appropriate headings and subheadings using the following hierarchy: BOLD CAPS, bold lower case, Plain text, Italics.
%     Tables should be in Word format and placed in the main text where the table is first cited.
%     Tables must be cited in the main text in numerical order.
%     Acknowledgments, Competing Interests, Funding and all other required statements.
%     Reference list.

\journaltitle{Journal Title Here}
\DOI{DOI HERE}
\copyrightyear{2023}
\pubyear{2023}
\access{Advance Access Publication Date: Day Month Year}
\appnotes{Paper}

\firstpage{1}

%\subtitle{Subject Section}

\title[Can GPT-3.5 Generate and Code Discharge Summaries?]{Can GPT-3.5 Generate and Code Discharge Summaries?}

\author[1,$\ast$]{Matúš Falis\ORCID{0009-0006-7649-6251} (MScR)}
\author[1]{Aryo Pradipta Gema\ORCID{0009-0007-1163-3531} (MScR)}
\author[2]{Hang Dong\ORCID{0000-0001-6828-6891} (PhD)}
\author[3]{Luke Daines\ORCID{0000-0003-0564-4000} (PhD)}
\author[4]{Siddharth Basetti\ORCID{0000-0002-7639-1859} (MBSS)}
\author[3]{Michael Holder\ORCID{0000-0001-9828-9153} (MMedSci)}
\author[5,6]{Rose S Penfold\ORCID{0000-0001-7023-7108} (BMBCh)}
\author[1]{Alexandra Birch\ORCID{0000-0002-9022-3405} (PhD)}
\author[7,8]{Beatrice Alex\ORCID{0000-0002-7279-1476} (PhD)}

\authormark{Falis et al.}

\address[1]{\orgdiv{School of Informatics}, \orgname{The University of Edinburgh}, \orgaddress{\state{Edinburgh}, \country{United Kingdom}}}
\address[2]{\orgdiv{Department of Computer Science}, \orgname{ University of Exeter}, \state{Exeter}, \country{United Kingdom}}
\address[3]{\orgdiv{Centre for Population Health Sciences, Usher Institute}, \orgname{The University of Edinburgh}, \state{Edinburgh}, \country{United Kingdom}}

\address[4]{\orgdiv{Department of Research, Development and Innovation}, \orgname{National Health Service Highland}, \state{Inverness}, \country{United Kingdom}}

\address[5]{\orgdiv{Ageing and Health, Usher Institute}, \orgname{The University of Edinburgh}, \state{Edinburgh}, \country{United Kingdom}}
\address[6]{\orgdiv{Advanced Care Research Centre}, \orgname{The University of Edinburgh}, \state{Edinburgh}, \country{United Kingdom}}
\address[7]{\orgdiv{Edinburgh Futures Institute}, \orgname{The University of Edinburgh}, \state{Edinburgh}, \country{United Kingdom}}
\address[8]{\orgdiv{School of Literatures, Languages and Cultures}, \orgname{The University of Edinburgh}, \state{Edinburgh}, \country{United Kingdom}}

\corresp[$\ast$]{Corresponding author: Mat\'{u}\v{s} Falis, School of Informatics, The University of Edinburgh, 10 Crichton Street, Edinburgh EH8 9AB, United Kingdom; \href{email:mfalis2@ed.ac.uk}{mfalis2@ed.ac.uk}}

% \received{Date}{0}{Year}
% \revised{Date}{0}{Year}
% \accepted{Date}{0}{Year}

% Leaving abstract empty here, and introduce it as a section below to comply with JAMIA's instruction
% https://academic.oup.com/jamia/pages/General_Instructions#:~:text=CHORUS%20initiative.-,Title,-page
\abstract{}
\counts{250 words in the abstract; 4,152 words in the main body; 4,428 words in the main body + Funding, Competing Interests, and Author Contribution sections; 4 figures (1 black and white, 3 colour); 4 tables; 34 references.}

\maketitle
\doublespacing

\clearpage

\section{\textbf{Abstract}}
\textbf{Objective:}
To investigate GPT-3.5 in generating and coding medical documents with ICD-10 codes for data augmentation on low-resources labels. \\
\textbf{Materials and Methods:}
Employing GPT-3.5 we generated and coded 9,606 discharge summaries based on lists of ICD-10 code descriptions of patients with infrequent (or generation) codes within the MIMIC-IV dataset. Combined with the baseline training set, this formed an augmented training set. Neural coding models were trained on baseline and augmented data and evaluated on a MIMIC-IV test set. We report micro- and macro-F1 scores on the full codeset, generation codes, and their families. Weak Hierarchical Confusion Matrices determined within-family and outside-of-family coding errors in the latter codesets. The coding performance of GPT-3.5 was evaluated on prompt-guided self-generated data and real MIMIC-IV data. Clinicians evaluated the clinical acceptability of the generated documents. \\
\textbf{Results:}
Data augmentation results in slightly lower overall model performance but improves performance for the generation candidate codes and their families, including one absent from the baseline training data. Augmented models display lower out-of-family error rates. GPT-3.5 identifies ICD-10 codes by their prompted descriptions, but underperforms on real data. Evaluators highlight the correctness of generated concepts while suffering in variety, supporting information, and narrative. \\
\textbf{Discussion and Conclusion:}
 While GPT-3.5 alone given our prompt setting is unsuitable for ICD-10 coding, it supports data augmentation for \mf{training} neural models. Augmentation positively affects generation code families but mainly benefits codes with existing examples. Augmentation reduces out-of-family errors. Documents generated by GPT-3.5 state prompted concepts correctly but lack variety, and authenticity in narratives. %\mf{Thus, given our prompt setting and experts' evaluation they are mainly suitable for data augmentation}.

%ABSTRACT CURRENTLY ON EXACTLY 250 WORDS!

\section{\textbf{BACKGROUND AND SIGNIFICANCE}}
\ac{LMTC} tasks in \ac{NLP} associate input documents with a set of output labels from a large label space, often hierarchically with a big-head long-tail distribution and data sparsity issues. Medical document coding is the task of assigning structured codes from a medical ontology -- \eg the \ac{ICD}\footnote{\url{https://www.who.int/standards/classifications/classification-of-diseases}} --  to clinical documents, a task performed by specially trained hospital staff. Coding consumes human resources that could be allocated to patient care. To ease this burden, research in machine learning and \ac{NLP} cast medical document coding as an \ac{LMTC} task \cite{dong2022automated}. 

In automatic ICD Coding, discharge summaries serve as input, yielding codes from a specified ICD version (\eg ICD-10-CM)\footnote{\url{https://www.cdc.gov/nchs/icd/icd-10-cm.htm}}. ICD coding faces distribution challenges mirroring other LMTC tasks. Few common conditions (\eg hypertension), contrast with many underrepresented or absent in corpora, such as MIMIC-IV \cite{johnson2023mimic}. Moreover, limited real-world data availability, often restricted for privacy reasons, compounds these challenges. However, modern deep learning ICD coding approaches (\eg CAML \cite{mullenbach2018explainable}, HLAN \cite{dong2021explainable}, RAC \cite{kim2021read}) are data-driven, and adversely affected by data sparsity unless explicitly designed to handle label under-representation. Techniques such as auxiliary information \cite{rios2018few, song2021generalized, ren2022hicu, wang2022hienet}, or data augmentation and synthesis \cite{falis2022horses, kim2022automatic, barros2022divide} attempt to mitigate these issues. ICD coding models with pre-trained encoders at best match the current state-of-the-art -- usually involving domain-specific versions of BERT \cite{afkanpour2022bert}. Large Language Models (LLMs) such as GPT-3 and its newer variants~\cite{instruct-tuning-gpt2022} (\eg GPT-3.5) or Large Language Model Meta AI~\citep[LLaMA;][]{touvron2023llama} have recently displayed state-of-the-art performance on several tasks with emerging capabilities \cite{zhao2023survey}. In the medical domain, notably Med-PaLM 2 \cite{singhal2022large} matching human performance in multiple-choice medical schools' exams. While using models such as GPT-3.5 is problematic with real discharge summaries due to privacy issues, these models have the potential to aid in generating synthetic discharge summaries for training local models.
%\footnote{In this work, we use GPT-3.5 to process clinical notes safely and mitigate the risks by turning off content monitoring from the API service provider, Microsoft Azure, by the MIMIC-III data usage requirement in \url{https://physionet.org/news/post/415}.}

Recently, Large Language Models (LLMs), notably GPT-3.5, have become a new standard for advanced NLP tasks, especially ones reliant on understanding natural language. These models retain and apply background knowledge observed during training, yet can also produce fluent but inaccurate information (known as hallucination) \cite{Ji2023hallucination-survey}. Writing and coding discharge summaries require extensive background knowledge making it of interest to explore an LLM's capability of reading, coding, and generating discharge summaries. An LLM capable of handling medical text could address the data sparsity issue by synthesising new data. This study aims to investigate the viability of GPT-3.5-generated medical documents for data augmentation in training local neural models and their credibility in clinical settings. We investigate GPT-3.5 given its performance in natural language understanding tasks requiring background knowledge, within an ethical experimental setting for clinical note data (by disabling content monitoring by the service provider). Furthermore, we explore GPT-3.5's ability to code real discharge summaries and self-generated text. 

Recent studies have prompted discussions on GPT's utility in medicine, including applications in medical chatbots \cite{lee2023benefits}, or radiology \cite{lecler2023revolutionizing}. \citet{yeung2023ai} compared the ChatGPT with a clinical GPT model \cite{kraljevic2022foresight} on generating patient vignettes. While research exists in generating data in low-resource settings in similar domains (\eg law \cite{ghosh2023dale}), to the best of our knowledge, GPT's performance in generating discharge summaries based on input conditions and its ability to perform ICD coding has not yet been reported.

\section*{\textbf{OBJECTIVE}}
\label{sec:obj}
This study aims to assess GPT-3.5's efficacy in the context of automated ICD-10 coding and investigate its viability as:

\begin{itemize}
    \item A data generator for ICD-10 coding, enhancing the training of local neural models by including GPT-3.5-generated discharge summaries, especially for rare labels;
    \item An automated ICD-coding classifier, either using synthetic text with explicit code descriptions or real data as prompts;
    \item A discharge summary generator focusing on producing clinically accurate and plausible synthetic data from expert perspectives.
\end{itemize}
\section*{\textbf{MATERIALS AND METHODS}}
\label{sec:mat}
We queried GPT-3.5 (gpt-3.5-turbo)\footnote{\url{https://platform.openai.com/docs/models/gpt-3-5}}, from hereon referred to as \emph{G}, through the OpenAI Python API\footnote{\url{https://platform.openai.com/docs/api-reference?lang=python}} to generate patient discharge summaries based on specific conditions and procedures represented by ICD-10-CM and ICD-10-PCS code descriptions from gold standard labels associated with MIMIC-IV discharge summaries (from the table \url{hosp/d_icd_diagnoses.csv.gz}). These label combinations were chosen to closely follow real scenarios to correspond to correlations between real labels.  Note that sharing MIMIC data via any online API is prohibited\footnote{\url{https://physionet.org/news/post/415}}\footnote{Our method was consulted with and approved by PhysioNet, as we merely use the descriptions of attached codes, which are not considered part of the dataset.}.

While various dataset splits have been proposed since the release of coded discharge summaries in MIMIC-IV, we specifically chose not to follow the one proposed by \citet{edin2023automated} due to its exclusion of the long tail, which contrasts with our aim of addressing this aspect through generation techniques. Instead, we adopted the dataset split proposed by \citet{nguyen2023mimic} (Table \ref{tab:dataset_split})that preserved the long-tail, aligning better with our focus. Nonetheless, we utilised the implementation of common ICD coding models produced by \citet{edin2023automated} for our analysis.

\begin{table}[]
\centering
% \begin{tabular}{l|rrr}
%                              & \multicolumn{1}{c}{Train} & \multicolumn{1}{c}{Dev} & \multicolumn{1}{c}{Test} \\ \hline
% \# Documents                 & 110,442                   & 4,017                   & 7,851                    \\
% \# Labels (Total)            & 1,784,304                 & 65,516                  & 124,518                  \\
% \# Labels (Unique)           & 25,230                    & 6,738                   & 9,159                    \\
% \# Labels (Unique Zero-Shot) & N/A                       & 291                     & 587                      \\
% \# Labels (Unique Few-Shot)  & 15,300                    & 919                     & 1,646                   
% \end{tabular}
\caption{MIMIC-IV Nguyen dataset split. Zero-Shot corresponds to labels absent from the training set. Few-Shot corresponds to labels appearing at least once but no more than five times in the training set.}
\begin{tabular}{lccc}
\toprule
                             & Train & Dev & Test \\
\midrule
\# Documents                 & 110,442   & 4,017  & 7,851   \\
\# Labels (Total)            & 1,784,304 & 65,516 & 124,518 \\
\# Labels (Unique)           & 25,230    & 6,738  & 9,159   \\
\# Labels (Unique Zero-Shot) & N/A       & 291    & 587     \\
\# Labels (Unique Few-Shot)  & 15,300    & 919    & 1,646   \\
\bottomrule
\end{tabular}
\label{tab:dataset_split}
\end{table}

\subsection{Label Selection}

\mf{Candidate source documents for generation were selected from MIMIC-IV based on label populations in our selected split. We identified codes common across training, validation, and test sets, choosing those appearing up to 5 times in the training set, resulting in 195 unique codes (compared to 15,353 unique few-shot codes in training). }

\mf{For these 195 codes we produced a list of codes belonging to their families (identified by the head of the code). Families with at least one relatively frequent code (population $>$ 100 in training) and at least one code exclusive to the test set (zero-shot) were retained leaving 16 families. The constraint of having at least one frequent code was included to explore the scenario where the model may predict a label due to its dominance in the population, and the increase in population of other labels potentially leading to more confusion. } 

\mf{From these 16 families we randomly selected 10 to generate from (E10, G43, H35, H81, S00, S02, S06, T82, T84, T85).} \mf{Of these, 114 codes had a population lower than 100 and are henceforth termed \emph{generation codes} (a list of codes is available in Supplementary Material 1).}

% We identified codes common across training, validation, and test sets, choosing those appearing up to 5 times in the training set, resulting in 195 unique codes (compared to 15,353 unique few-shot codes in training). 

% For these 195 codes we produced a list of codes belonging to their families (identified by the head of the code). Families with at least one relatively frequent code (population $>$ 100 in training) and at least one code exclusive to the test set (zero-shot) were retained leaving 16 families. 

\subsection{Preparation of Samples for Generation}
As there exist co-relations among labels (\eg different complications of type 1 diabetes, cancer co-relating with the presence of chemotherapy), rather than creating random combinations of ICD codes we opted to work our way back from existing real scenarios.

In Nguyen's MIMIC-IV training set, we found documents with the 98 relevant few-shot codes (the 16 zero-shot were by-definition absent). Some documents contained multiple relevant codes. We have collected documents for each of the relevant codes and cloned them to bring their population up to 100. To increase variety, we have randomly dropped up to 5 of the assigned non-relevant labels within clones to create the new set of labels for generation (referred to as the silver standard). 

We identified documents containing the siblings of the 16 zero-shot labels. A silver standard set was created for each of these documents substituting the sibling code with the zero-shot code (similar to the zero-shot approach in \cite{falis2022horses}). If multiple siblings were present, a random one was replaced with the zero-shot code. This resulted in 9,606 input sets of labels -- 6,779 unique and 2,827 duplicated.

\subsection{Generation}
Natural Language Generation is the task of producing natural language text based on a set of input data. We used the model ``gpt-3.5-turbo-0613'' given its wide recognition in the field of LLMs, relative cost-effectiveness, and time efficiency (compared to gpt-4).  We utilised a temperature (parameter in the 0-1 range controlling randomness) of 0 to produce deterministic outputs. We set the temperature to 0.1 for duplicates, allowing output variation.

Within the prompt (see Supplementary Material 2), we specified to write a discharge summary for a patient with a list of standard descriptions of their conditions and procedures based on our silver standard. We added further specifications:

\begin{figure}
    \centering
    \includegraphics[width=15cm]{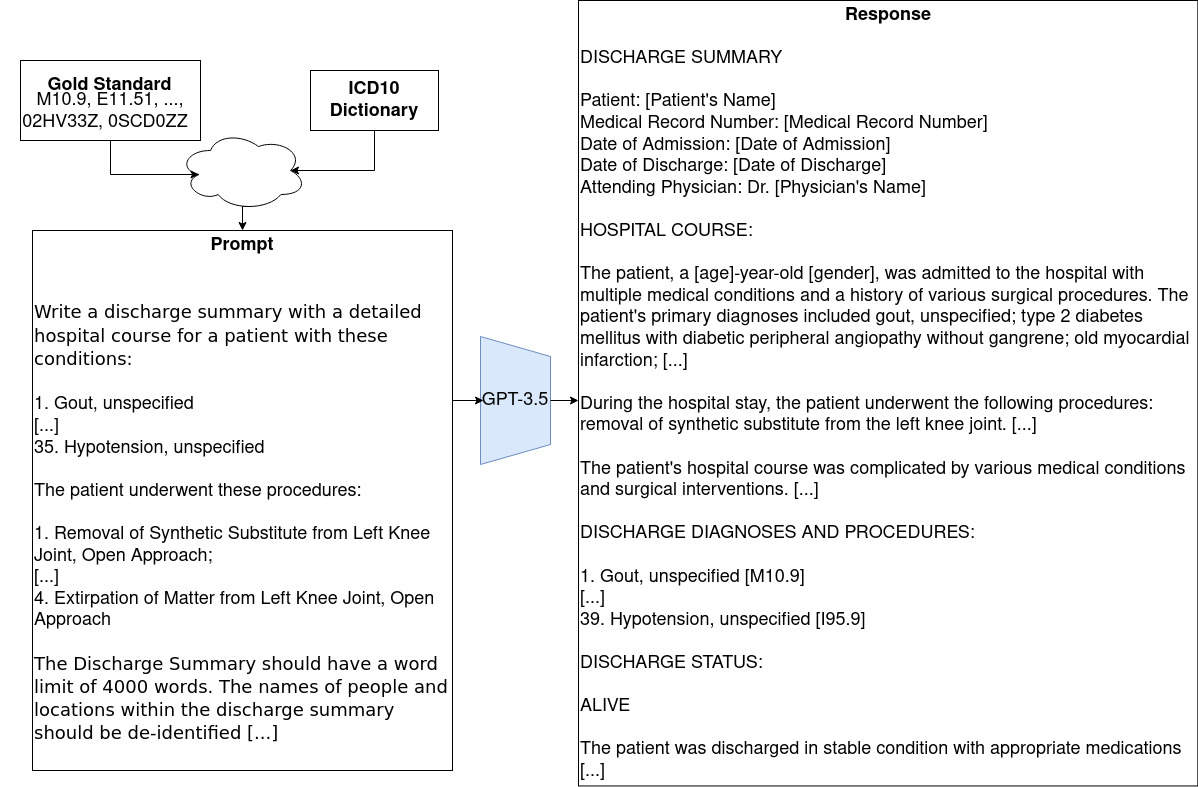}
    \caption{An example generation of a synthetic discharge via GPT-3.5}
    \label{fig:example_gen}
\end{figure}

\begin{itemize}
    \item Length of up to 4,000 words (following the maximum cutoff point in previous work \cite{edin2023automated}). The overall input and output token restriction of GPT-3.5 is 4,096 (inclusive of the prompt);
    \item Inclusion of social and family history;
    \item Anonymisation was required for personal and location data (due to uncertainty of the anonymity of the data used in training of GPT-3.5), maintaining numeric information when relevant despite potential removal in pre-processing;
    \item Explicit ICD code mentions within the main text were to be avoided to prevent model association or potential errors;
    \item Clear numeric values were preferred rather than ranges, especially for time-related codes;
    \item Providing a specific concept for codes involving the umbrella term ``other'' encompassing a range of conditions;
    \item Omission of the keyword ``unspecified'' present in standard descriptions  opting for a more natural means of expression;
    \item Coding of the discharge summary was to be positioned at the end in a regular pattern (codes in square brackets) for model coding assessment.
\end{itemize}

An example of the generation process can be seen in Figure \ref{fig:example_gen}.

The generated documents were processed to find the discharge diagnoses and extract the assigned codes with a regular expression.  In total, we have generated 9,606 synthetic training documents. We removed all mentions of ICD-10 codes and pre-processed the documents the same as the baseline data. Then we merged these documents with the \emph{baseline} training set (110,442 MIMIC-IV documents), forming the \emph{augmented} training set (120,048 documents). The baseline and augmented settings used the same validation and test sets with 4,017 and 7,851 real documents, respectively.

\begin{figure}
\begin{minipage}{.55\linewidth}

\centering
\subfloat[]{\label{fig:words_per_label}\includegraphics[scale=0.27]{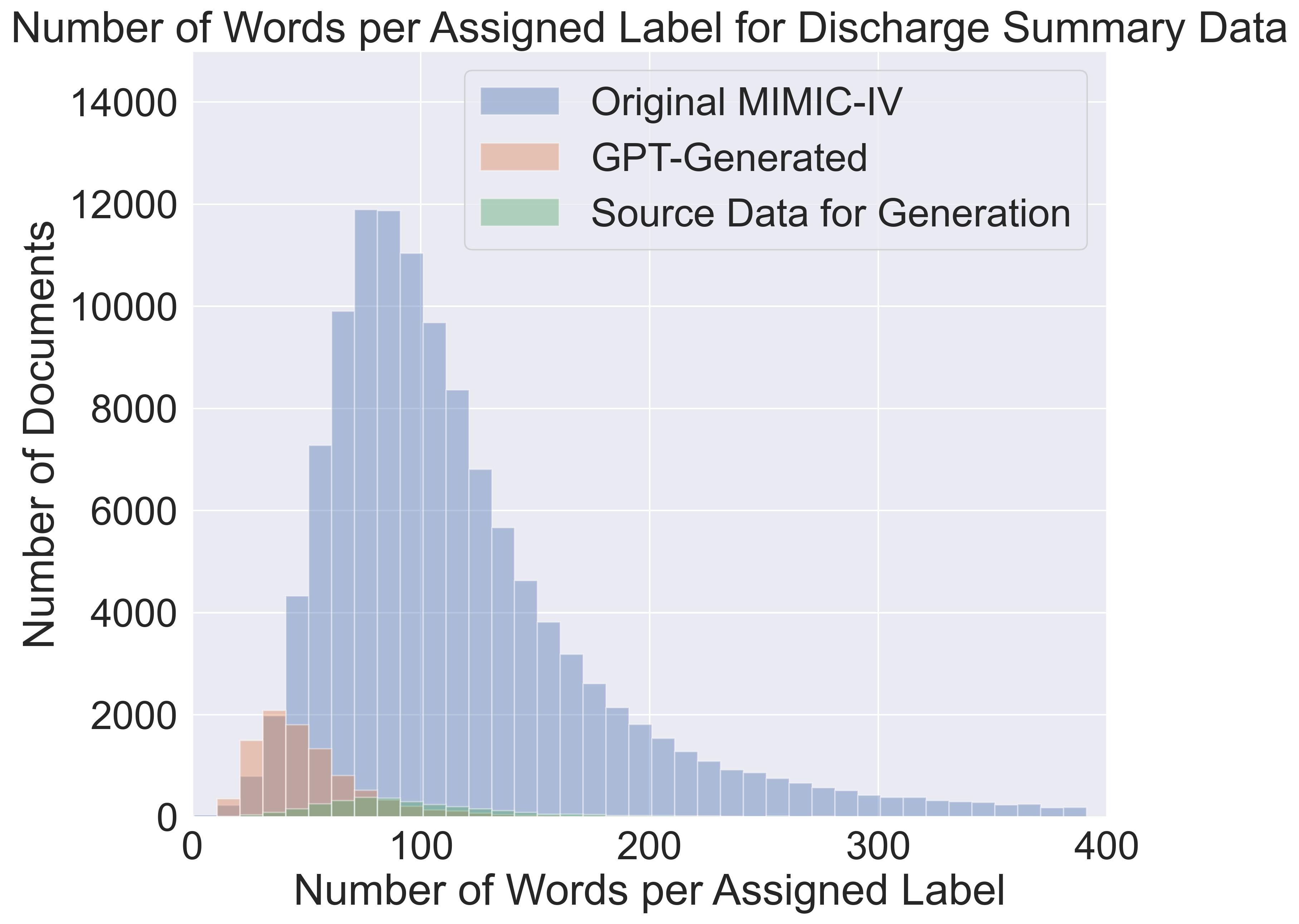}}
\end{minipage}
\begin{minipage}{.55\linewidth}
\centering
\subfloat[]{\label{fig:words_in_doc}\includegraphics[scale=.27]{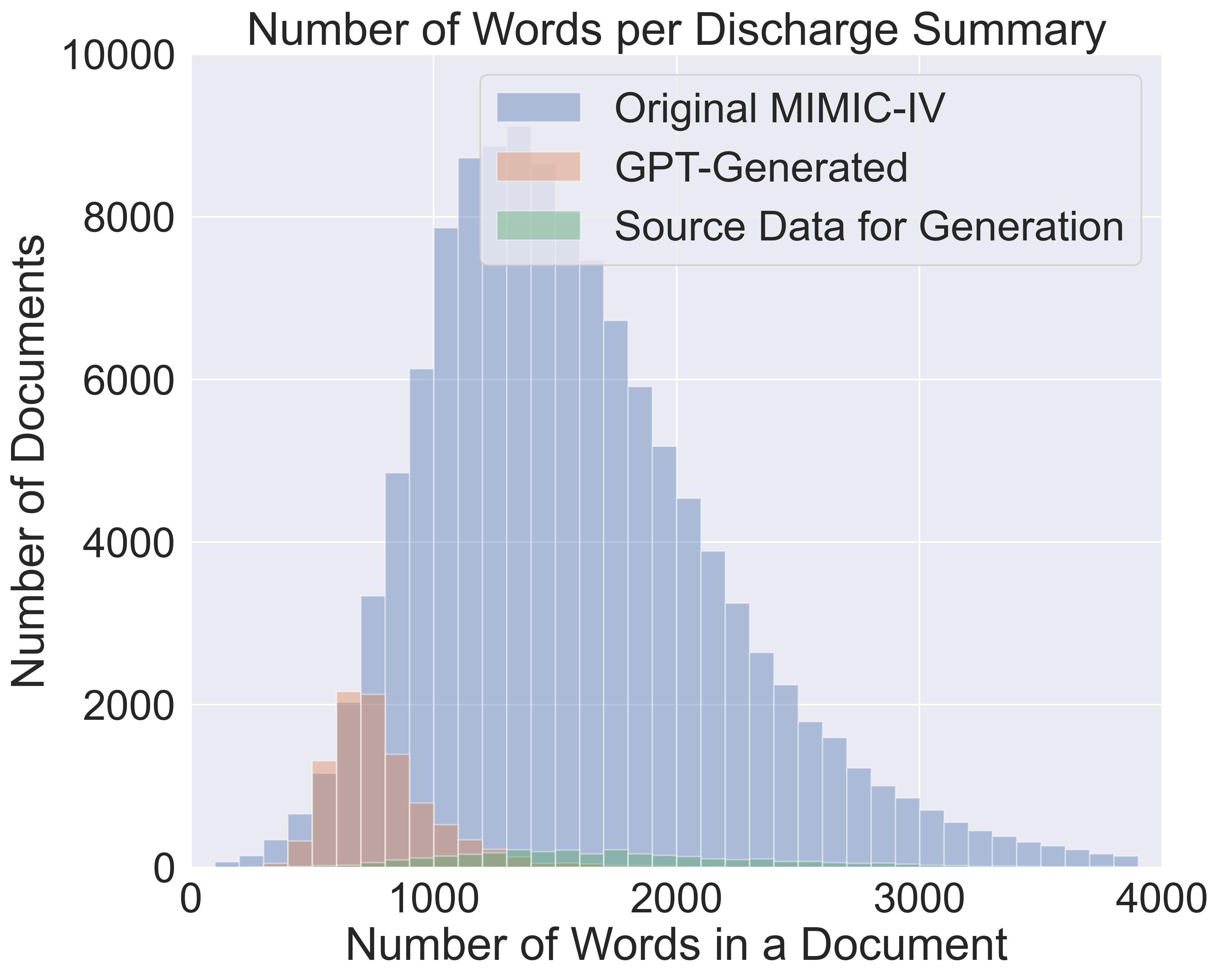}}
\end{minipage}%
\par\medskip

\centering
\subfloat[]{\label{fig:targets_in_doc}\includegraphics[scale=.27]{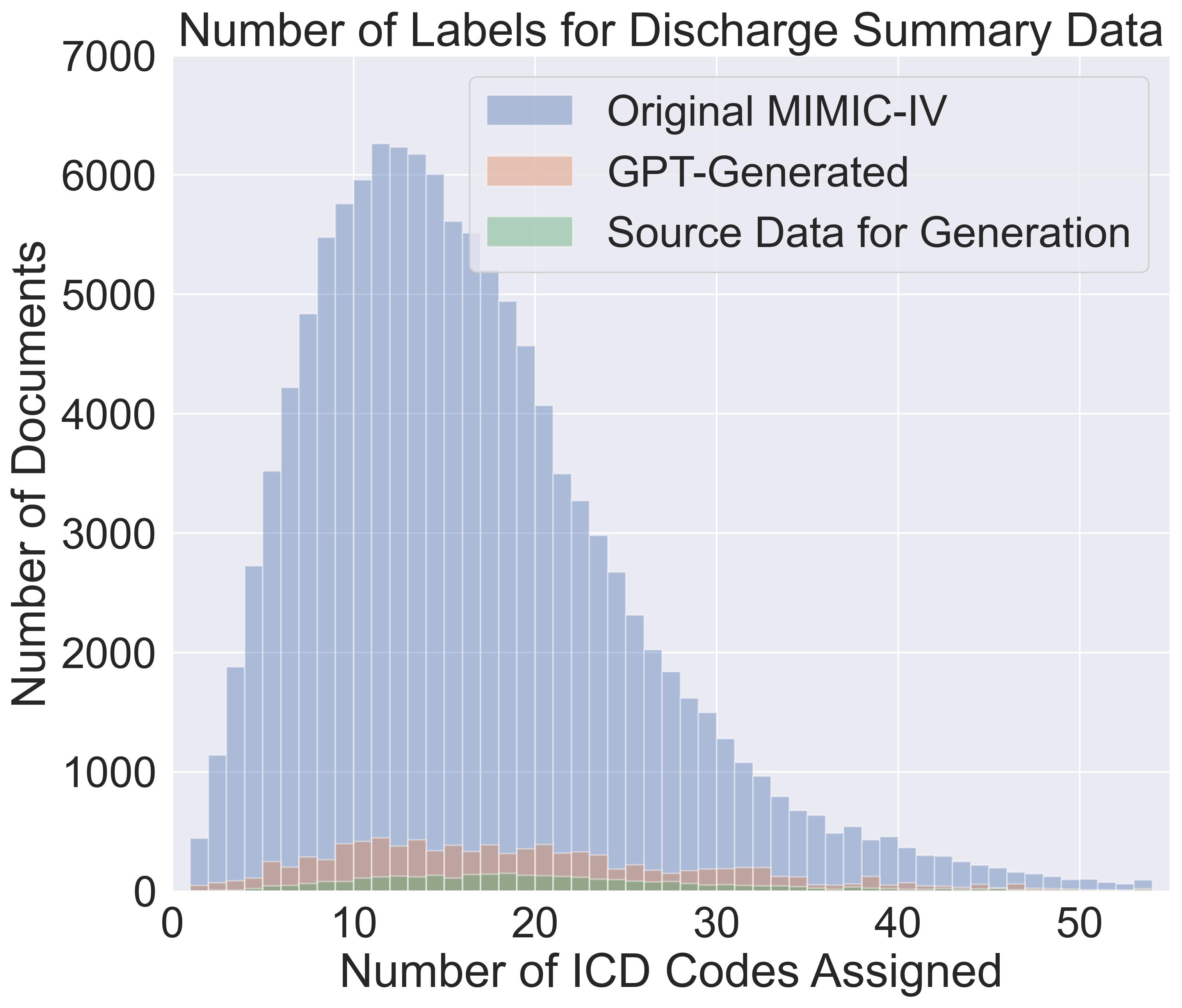}}

\caption{A comparison between the discharge summary data in MIMIC-IV, seed MIMIC discharge summaries for generation (the source data), and the generated discharge summaries. Subfigures \ref{fig:words_per_label} and \ref{fig:words_in_doc} focus on the number of words in documents, indicating that the GPT-generated data generally contains fewer words overall and per assigned label compared to the real data from MIMIC-IV.  Subfigure \ref{fig:targets_in_doc}  demonstrates that, although there's a variance in document size, the distribution of the number of labels per document remains relatively similar across the datasets.}
\label{fig:words_stats}
\end{figure}

Analysis between the generated text, and MIMIC-IV ICD-10-coded discharge summaries (the entire dataset, and the subset used for source label sets used in generation) showed differences in word count (per label and overall -- Figures \ref{fig:words_per_label} and \ref{fig:words_in_doc}) between synthetic and real data while retaining similar distributions of label counts (Figure \ref{fig:targets_in_doc}). The synthetic discharge summaries tend to be shorter which could be the result of GPT-3.5's 4,096 token limit. The synthetic discharge summaries have been accepted for publication and will be and will be made available by PhysioNet.

\subsection{Local Neural Models}
Most recent \ac{LMTC} neural architectures are encoder-decoder models whose encoder processes the input text to generate a latent representation. 

Architectures using a non-BERT-like encoder (\eg in CAML \cite{mullenbach2018explainable}, LAAT \cite{vu2020label}, or Multi-Res CNN \cite{li2020icd}) utilise non-contextual (\eg Word2Vec \cite{mikolov2013distributed}) word embeddings, while BERT(\cite{devlin2018bert})-like encoders (\eg in PLM-ICD\cite{huang2022plm}) enable contextual token representation.
The decoder determines a probability for each label based on the latent representation. A probability threshold determines positive predictions.

\subsection{Evaluation}
We have conducted four evaluation rounds:
\begin{itemize}
    \item \textbf{Local Neural Model Evaluation:} Assessing CAML, LAAT, and Multi-Res CNN models' performance on Nguyen's test set. Models were trained either solely on Nguyen's training set or enhanced with data generated by \emph{G} (the augmented training set).
    \item \textbf{GPT's coding on real data:} Evaluation of \emph{G}'s coding ability on MIMIC-IV using Nguyen's test set;
    \item \textbf{GPT's coding on GPT-generated data:} Evaluation of \emph{G}'s coding ability on generated documents (with provided code descriptions in the prompt)
    \item \textbf{Acceptability of Generated Data in Clinical Practice:} Reviewing \emph{G}-generated discharge summaries by clinical professionals to gauge their suitability in clinical settings.
\end{itemize}
\subsubsection{Local Neural Model Evaluation}
We used the codebase and training procedure of \citet{edin2023automated} to train and evaluate CAML, LAAT, and Multi-Res CNN on the ICD coding task using Nguyen's split. Each model is trained with 20 epochs on the training set. Throughout the training, we evaluated the model on the validation set, selecting the model with the highest mean average precision (MAP) from each run as the final model. Subsequently, this final model underwent evaluation on the test set. Additionally, we explored PLM-ICD, considered a state-of-the-art model for this task. However, its performance on the baseline and augmented data was notably lower than previously reported. This model has been reported to be unstable during training, which depends on random seeds. We decided against tuning the random seed, and only report the performance for CAML, LAAT, and Multi-Res CNN.

The models' test performance was evaluated using standard information retrieval metrics commonly used in LMTC tasks -- micro- and macro-averaged Precision, Recall, and F1 scores.

Micro-averaging assigns equal weight to each prediction, favouring high-population classes (\eg hypertension). Macro-averaging, in contrast, computes the performance for each unique label and averages across the label space, giving each label's average result equal weight regardless of their population. This highlights poor performance in less common classes.  Our primary evaluation metrics common with the majority of previous work are micro-F1 and macro-F1 scores. Metrics are further explained in Supplementary Material 3.

\subsubsection{GPT's coding on real clinical notes}

\begin{figure}
    \centering
    \includegraphics[width=\textwidth]{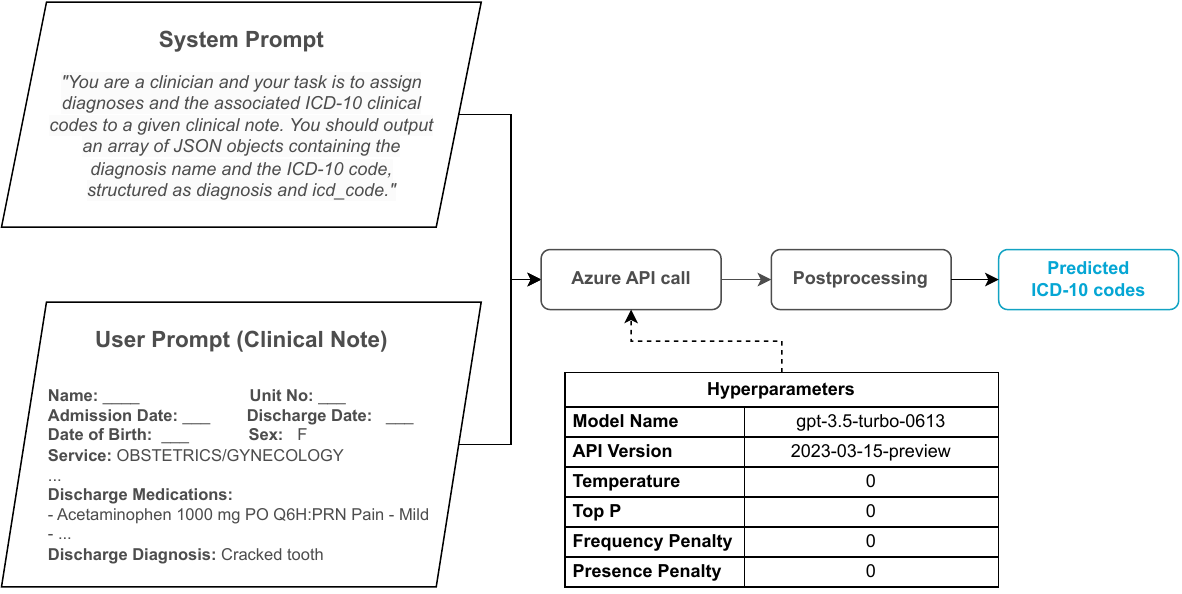}
    \caption{The workflow of the GPT-3.5 prediction. We used Azure AI Services API to query GPT-3.5 and we employed a postprocessing step to extract the predicted diagnoses and ICD-10 codes for each clinical note.}
    \label{fig:chatgpt_api_call}
\end{figure}

We used the Azure AI Services API\footnote{https://azure.microsoft.com/en-gb/products/ai-services} to test GPT's ability to assign diagnosis codes based on real clinical notes.  
The API returns a free text response which we further processed to retrieve the predicted ICD-10 codes.
% Postprocessing
In postprocessing, we verify the response format.
For correctly structured arrays of JSON objects we simply extract the predictions. For incorrectly structured outputs, we employ a regular expression pattern to extract all diagnoses and ICD code pairs.
The result is a list of predicted diagnoses and corresponding ICD-10 codes for each clinical note.
Figure~\ref{fig:chatgpt_api_call} illustrates the API call workflow.

% Azure AI API
For reproducibility, we specified the model version and the API version as \textit{``gpt-3.5-turbo-0613''} and \textit{``2023-03-15-preview''}, respectively.
%% All params to ensure deterministic response
All parameters were set to zero for deterministic responses from \emph{G}, including temperature, top P, frequency penalty, and presence penalty.
%% System prompt
The system prompt directed \emph{G} to act as a clinician assigning ICD-10 diagnosis codes to clinical notes, specifying the expected output format as JSON objects with keys \textit{``diagnosis''} and \textit{``icd\_code''}.
Refer to Figure \ref{fig:chatgpt_api_call} for all hyperparameter details.
Our code implementation can be found in our Github repository~\footnote{\href{https://github.com/EdinburghClinicalNLP/chatgpt_icd_coding}{https://github.com/EdinburghClinicalNLP/chatgpt\_icd\_coding}}.

% %% Table for API params: model name, version, max tokens, temperature, top_p, freq penalty, presence penalty
% \begin{table}
%     \centering
%     \begin{tabular}{lp{0.6\linewidth}}
%     \toprule
%     Parameter & Value \\
%     \midrule
%     Model Name        & gpt-3.5-turbo-0613 \\
%     API Version       & 2023-03-15-preview \\
%     Temperature       & 0 \\
%     Top P             & 0 \\
%     Frequency Penalty & 0 \\
%     Presence Penalty  & 0 \\
%     System Prompt     & You are a clinician and your task is to assign diagnoses and the associated ICD-10 clinical codes to a given clinical note. You should output an array of JSON objects containing the diagnosis name and the ICD-10 code, structured as diagnosis and icd\_code. \\
%     \bottomrule
%     \end{tabular}
%     \caption{Azure API call hyperparameters.}
%     \label{tab:api_hyperparameters}
% \end{table}

% Opting out of human review
We opted out of human review of the data for two reasons.
Firstly, the terms of the data use agreement of MIMIC-IV~\footnote{https://physionet.org/news/post/415} did not grant us the authority to permit a third party to process the data for abuse detection.
Secondly, we assessed the likelihood of harmful misuse to be low given the sensitive nature of the clinical notes.

In our evaluation of GPT's performance, we have also employed hierarchical evaluation techniques -- set-based hierarchical evaluation ~\cite{kosmopoulos2015evaluation} and Count-Preserving Hierarchical Evaluation (CoPHE)~\cite{falis2021cophe}. These metrics award partial credit to mispredicted labels by extending prediction and gold standard sets with their ancestor labels.
While set-based evaluation ancestor labels track only the presence of descendants, in CoPHE, ancestor labels link to the count of descendants, penalising over- and under-predictions within code families.

Comparing set-based and CoPHE results helps to evaluate the model's tendency to over-/under-predict. A lower CoPHE score indicates this phenomenon. See Supplementary Materials 4 for details on calculating hierarchical scores.

We utilise macro-averaged metrics from Weak Hierarchical Confusion Matrices (WHCM) \cite{falis2022horses} to summarise in-family versus out-of-family (OOF) prediction errors. These metrics are chosen to explore how expanding the population of codes within code families to a minimum of 100 instances impacts within-family performance.  Within-family errors involve false positives that align with false negatives within the same family in the gold standard. On the other hand, an OOF error for a false negative in the gold standard lacks a false positive within the prediction set from the same family to match. Our primary goal in generating synthetic data is to reduce OOF errors, enhancing true positive predictions or ensuring mispredictions occur within-family.

\subsubsection{GPT-3.5's Coding on Synthetic Data}
The prompt asked \emph{G} to code the conditions and procedures mentioned in the document it generated. This experiment tested \emph{G}'s ability to assign ICD-10 codes to concepts presented in their standard descriptions. Alongside this, the prompt required creating a patient's social and family history, which might have led to the model introducing new conditions like substance abuse and potentially coding them, despite not being part of the initial prompt.

\subsubsection{Acceptability of Generated Data in Clinical Practice}
Four clinical professionals (co-authors SB, LD, MH, and RSP) assessed the quality of the generated data. As the data was generated based on labels associated with MIMIC-IV discharge summaries, this evaluation included both synthetic discharge summaries generated by \emph{G} and discharge summaries from MIMIC-IV. The clinicians were presented with 20 discharge summaries -- 10 synthetic and 10 real (based on whose adjusted gold standard the synthetic ones were generated). 

Each discharge summary was assessed for:

\begin{itemize}
\item Correctness -- accuracy in describing patient conditions and procedures;
\item Informativeness -- clarity and sense in supporting information (\eg test results, medication suggestions);
\item Authenticity (patient) -- whether such a patient could exist;
\item Authenticity (clinical scenario) -- whether the hospital course was plausible as reported;

\item Acceptability -- suitability of the document for clinical use.
\end{itemize}

\begin{figure}
    \centering
    \includegraphics[width=\textwidth]{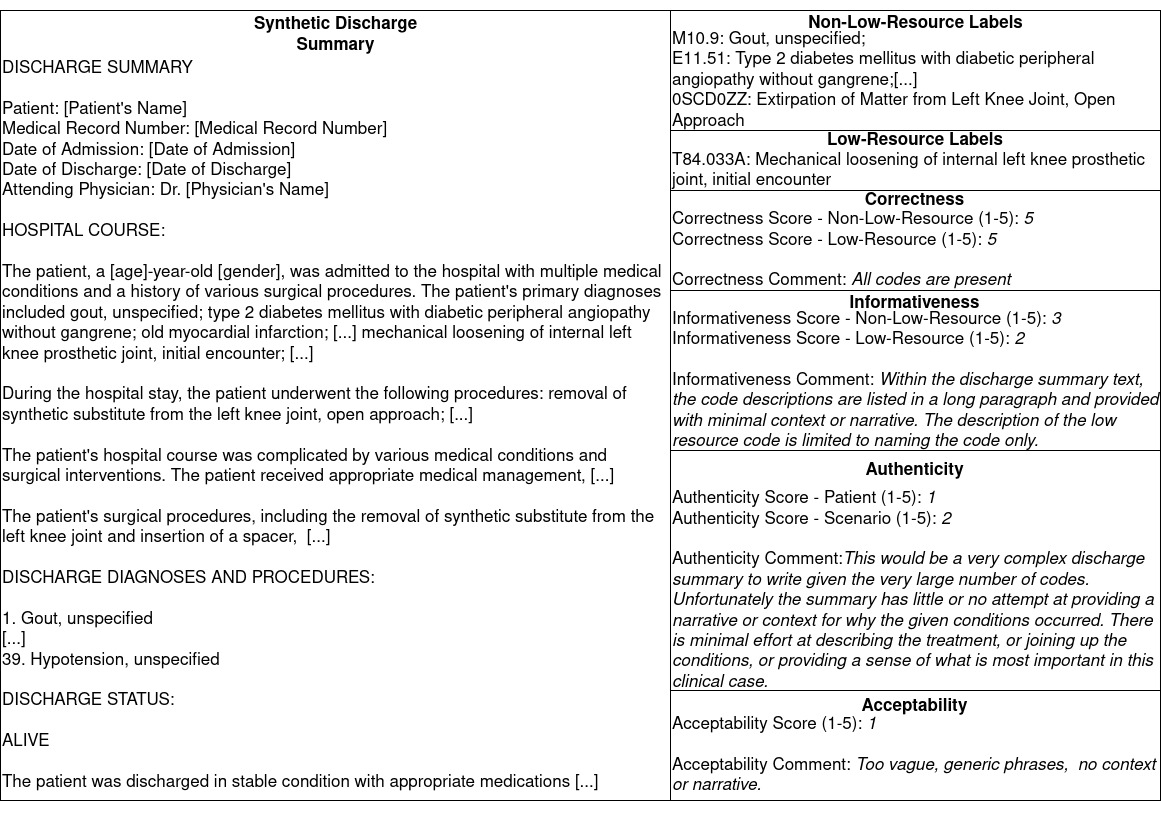}
    \caption{An example evaluation of a synthetic discharge summary by a clinical expert.}
    \label{fig:example_eval}
\end{figure}

Additionally, they separately evaluated correctness and informativeness for both non-low-resource and low-resource labels to gauge \emph{G}'s ability to generate low-resource data. Scores from 1 (failure to perform) to 5 (perfect performance) were assigned to each metric, with accompanying comments justifying the score. An example evaluation by a clinician can be seen in Figure \ref{fig:example_eval}.

\section{\textbf{RESULTS}}

\subsection{Local Neural Model Evaluation}

We assessed performance across three code subsets: the entire codeset in MIMIC-IV ($overall$), a restricted generation set ($f_{(\text{gen})}$) containing only the 114 low-population candidate generation labels, and a set of codes from the code families present in $f_{(\text{gen})}$ ($f$). Results are shown in Table \ref{tab:augment_f1_performance}. Baseline results for CAML and LAAT align closely with prior findings by \citet{nguyen2023mimic} for $overall$ metrics. While \citet{nguyen2023mimic} did not report on Multi-Res CNN, its performance trends were similar to CAML and LAAT in comparison to \citet{edin2023automated} on non-filtered codesets. LAAT excels in micro-F1, Multi-Res CNN leads in macro-F1, and CAML generally lags behind.  Baseline models outperform augmented ones in $overall$ micro-F1, a common observation when enhancing lower-resource label performance.  Nonetheless, macro-F1 scores improved for two out of three models within the $overall$ codeset and for all models on $f$ and $f_{(\text{gen})}$. Multi-Res CNN and CAML macro-F1 scores display sizable relative improvement (26\% and 78\%  respectively) in $f_{(\text{gen})}$.

Augmented models performed on par with or outperformed baseline models in micro-F1 scores for $f$ and $f_{(\text{gen})}$. Augmented Multi-Res CNN outperforms its baseline in micro-F1 for both $f$ and $f_{(\text{gen})}$, indicating benefits for the code family from augmenting less-populous members. Augmented LAAT shows improvement in macro-F1 in both $f$ and $f_{(\text{gen})}$ but lags in micro-F1.  \mf{LAAT's performance may have been biased towards high-population classes and the augmentation's boosting of low-frequency classes (misrepresenting their frequency) may have introduced confusion}.  \mf{Apart from having a recurrent encoder (Bi-LSTM), the LAAT model employed in this experiment is about twice the size of the Multi-Res CNN (21.9M versus 11.9M parameters). This added model complexity may have enabled better performance on already frequent labels, but increased the need for more examples of lower-resource labels.}

Comparing within-family and out-of-family errors (Table \ref{tab:augment_f1_performance}), augmented models generally exhibit fewer out-of-family errors on $f$ and $f_{(\text{gen})}$. An exception to this is augmented Multi-Res CNN in $f_{(\text{gen})}$, whose slight increase in OOF came with a sizable reduction in within-family error.

Unlike baseline models, augmented ones occasionally predicted codes absent from the original data, although incorrectly, except for one correctly predicted code (S02.63XA) by a Multi-Res CNN model trained on augmented data. While consistent enhancement in zero-shot code performance was not achieved through augmentation, the potential for improvement is evident.

\begin{table}[]
\centering
\caption{A comparison between local neural network models (MRCNN stands for Multi-Res CNN) trained on baseline ($base$) and augmented ($aug$) training sets is evaluated using micro- and macro-averaged $F1$ scores ($mi$ and $ma$ respectively)  on three codesets -- $overall$ on all codes present in MIMIC-IV; $f$ comprising all codes within the families we chose for generation; and $f_{(\text{gen})}$ corresponding to candidate codes used in generation with a population of at most 100 in the training set. The highest score in each metric for each model pair (baseline versus augmented) is highlighted in \textbf{bold}. Weak Hierarchical Confusion Matrix (WHCM) error rates are produced for codesets $f$ and $f_{(\text{gen})}$. Performance on the common test set is reported using the macro-averaged proportion of errors that were Out-of-Family (OOF) and within-family (IF). The best (lowest) error rate for each error type for each model pair (baseline versus augmented) is presented in bold.}
\begin{tabular}{lcccccccccc}
    \toprule
    \multirow{2}{*}{Experiment} & \multicolumn{6}{c}{F1 $\uparrow$} & \multicolumn{4}{c}{WHCM error $\downarrow$} \\
    \cmidrule(r){2-7} \cmidrule(l){8-11}
         & $mi_{overall}$    & $ma_{overall}$    & $mi_{f}$          & $ma_{f}$         &  $mi_{f(\text{gen})}$    & $ma_{f(\text{gen})}$      & $OOF_f$         & $IF_f$           & $OOF_{f(\text{gen})}$   & $IF_{f(\text{gen})}$   \\
    \midrule
    CAML$_{base}$  & \textbf{53.65}  & 3.87   & \textbf{38.43}  & 3.03  & 17.41  & 6.64    & 66.53           & 25.05 & 83.81 & 9.83            \\
    CAML$_{aug}$   & 53.54 & \textbf{3.90} & 38.41 & \textbf{3.78} & \textbf{20.68} & \textbf{11.86} & \textbf{65.98} & \textbf{23.77} & \textbf{79.79} & \textbf{9.17} \\
    \midrule
    LAAT$_{base}$  & \textbf{57.29}  & \textbf{6.18}   & \textbf{43.59}  & 4.96  & \textbf{26.79}  & 14.48   & 58.57           & \textbf{28.35} & 74.03 & 12.20           \\
    LAAT$_{aug}$   & 57.18 & 6.09 & 43.36 & \textbf{5.38} & 25.70 & \textbf{14.98} & \textbf{55.93} & 29.78 & \textbf{73.65} & \textbf{11.98} \\
    \midrule
    MRCNN$_{base}$ & \textbf{55.66}  & 6.40   & 40.16  & 5.04  & 26.80  & 13.92   & 52.72           & 32.41 & \textbf{69.68} & 15.24           \\
    MRCNN$_{aug}$  & 54.69 & \textbf{6.46} & \textbf{42.69} & \textbf{5.85} & \textbf{30.39} & \textbf{17.68} & \textbf{49.65} & \textbf{32.36} & 70.41 & \textbf{10.22} \\
    \bottomrule
\end{tabular}
\label{tab:augment_f1_performance}
\end{table}
\subsection{GPT's coding ability on real and synthetic data}
We examined \emph{G}'s ability to code real MIMIC-IV documents and generate coded documents with explicit code descriptions in the prompt. The results (Table \ref{tab:gpt-coding}) show that the performance on prompt-guided self-generated (synthetic) data resembles that of local models on the MIMIC-IV test set, not surpassing it. Hierarchical metrics show higher precision, recall, and consequently, F1-score in CoPHE compared to set-based hierarchical evaluation indicating errors coming from within-family misprediction, rather than incorrectly estimating the number of expected labels. 

However, the performance on the MIMIC-IV test set is notably low, especially in precision. The improvement in the precision from leaf-only results to hierarchical is minimal. This implies that incorrect predictions were more likely to be out-of-family. Moreover, results on CoPHE are lower than on the set-based hierarchical evaluation indicating a tendency of the model to over-/under-predict within the scope of the family -- an issue previously reported in local ICD coding models \cite{falis2021cophe}, and present in the reported hierarchical results for baseline LAAT. 

These results demonstrate that \emph{G} can identify ICD-10 codes \mf{in self-generated mentions} based on provided descriptions if presented within the prompts. Its performance when tasked with standard ICD coding without explicitly identified concepts or non-standard surface forms of the concepts significantly deteriorates.

\begin{table}[]
\centering
\caption{Results of GPT-3.5's coding ability on real and self-generated data.}
\begin{tabular}{lccccccccc}
\toprule
\multirow{2}{*}{Evaluation Set} & \multicolumn{3}{c}{Leaf-Only} & \multicolumn{3}{c}{Set-Based} & \multicolumn{3}{c}{CoPHE} \\
\cmidrule(lr){2-4} \cmidrule(lr){5-7} \cmidrule(lr){8-10}
                                & P      & R      & F1     & P      & R      & F1     & P      & R      & F1     \\
\midrule
GPT-3.5 Real                            & 9.46   & 33.51  & 14.76  & 10.59  & 44.87  & 17.13  & 10.30  & 44.33  & 16.72  \\
GPT-3.5 Synthetic                       & 59.06  & 40.72  & 48.20  & 66.46  & 41.32  & 50.96  & 67.20  & 41.55  & 51.35  \\
Best Baseline (LAAT) Real     & 60.42  & 54.46  & 57.29  &  61.28
     &  54.50
      &  57.68
      &  60.84      & 54.33       &   57.39     \\
\bottomrule
\end{tabular}
\label{tab:gpt-coding}

\end{table}

\subsection{Acceptability of Generated Data in Clinical Practice}
\label{sec:res}

We calculated the inter-evaluator agreement for the 7 metrics using Fleiss' kappa ($\kappa$) \cite{fleiss1971measuring}. As $\kappa$ is designed for categorical variables and does not fully capture ordinal scores, also produced the mean scores for each metric ($\mu$). The results are presented in Table \ref{tab:human_eval}. 

The evaluators' agreement was poor ($\kappa<0$) in examples from MIMIC-IV for Correctness and Informativeness of non-low-resource codes, and the Acceptability of the discharge summaries. For the other metrics, a $\kappa>0$ was reached but never exceeded 0.4 (lower than moderate agreement). All mean scores are higher than 4. Hence, while the clinicians disagreed on the exact scores, they rated real discharge summaries positively. The disagreement may be due to clinicians being UK-based with significant differences in reporting style within the UK and the US (where MIMIC-IV is from).

For GPT-generated summaries, slight agreement was seen in acceptability, and fair agreement in the correctness of low-resource labels. All other metrics had poor agreement. Both correctness metrics scored above 4, with low-resource correctness surpassing 4.5—an encouraging outcome for our primary goal of generating low-resource code data.  Mean informativeness in the low-resource scenario and authenticity of scores were at least 3. Once again, performance on the low-resource codes exceeded non-low-resource codes. Other metrics had $\mu$ scores above 2. Informativeness and authenticity for non-low-resource codes had a poor agreement ($\kappa<0$), while acceptability had some agreement with the lowest mean score of 2.225.

\begin{table}[h]
\centering
\caption{Evaluator agreement ($\kappa$) and mean scores ($\mu$) for samples from MIMIC-IV (real), versus GPT-generated (synthetic) data.}
\begin{tabular}{lcccc}
\toprule
Metrics & $\kappa_{\text{real}}$ & $\mu_{\text{real}}$ & $\kappa_{\text{synthetic}}$ & $\mu_{\text{synthetic}}$ \\
\midrule
Correctness -- Non-Low-Resource     & -0.386 & 4.175 & -0.163 & 4.375 \\
Correctness -- Low-Resource         & 0.043  & 4.350 & 0.206  & 4.525 \\
\midrule
Informativeness -- Non-Low-Resource & -0.155 & 4.550 & -0.220 & 2.775 \\
Informativeness -- Low-Resource     & 0.241  & 4.675 & -0.277 & 3.000 \\
\midrule
Authenticity -- Patient             & 0.340  & 4.750 & -0.078 & 3.150 \\
Authenticity -- Scenario            & 0.373  & 4.775 & -0.333 & 2.250 \\
\midrule
Acceptability                       & -0.056 & 4.550 & 0.035  & 2.225 \\                              
\bottomrule
\end{tabular}
\label{tab:human_eval}
\end{table}

% ChatGPT failed to create natural-looking clinical notes
% \subsubsection{ChatGPT failed to create natural-looking clinical notes}
While \emph{G} generally produces correct notes, the clinical evaluators have identified several challenges in the generation of natural-looking clinical notes:

%% ChatGPT being verbatim, listing all diagnoses mentioned in the prompt
%% Real clinical notes may miss some diagnoses (less important ones) despite they can be inferred from other supporting information
\subsubsection{GPT-3.5 tends to do verbatim reproductions of the prompted diagnoses list}

\emph{G} tends to copy all concepts mentioned in the prompt when generating a clinical note.
While instruction-following is a desirable behaviour, excluding non-crucial details is essential when generating a natural-looking clinical note. 
Real clinical notes often omit irrelevant and less critical findings for brevity, particularly if the information is inferrable from surrounding contexts such as medications and treatments.
For instance, \emph{G} unnecessarily noted a normal BMI.

%% ChatGPT being verbatim, worded the diagnoses in unnatural ways (e.g. "anaemia, which was unspecified")
\subsubsection{GPT-3.5 may phrase diagnoses in an unnatural manner}

\emph{G} tends to use an overly technical and unnatural style when specifying diagnoses.
For instance, \emph{G} mentioned \textit{``anaemia, which was unspecified.''} in the generated clinical note as it was prompted with \textit{``D64.9: Anemia, unspecified''}.
\emph{G} also occasionally introduces vague phrases (\eg \textit{``geriatric team provided supportive care, including behavioural interventions and medication management''}) without further detail.
% This unnatural style easily differentiates synthetic clinical notes from real ones.
This contrasts with the more streamlined language of real clinical notes.

%% Missing necessary supporting information
\subsubsection{GPT-3.5 lacks details when introducing supporting information}

\emph{G} tends to introduce crucial supporting information without sufficient details.
For instance, \emph{G} mentioned \textit{``Following a traumatic event''} without further specification of the mentioned traumatic event, which is unacceptable in the clinical setting.
% \emph{G} also mentioned \emph{pleural intervention} without additional detail or rationale.
This omission limits the overall informativeness of the patient's medical context, potentially hindering the notes' usability for a comprehensive view.

%% or straight-up spurious supporting information
\subsubsection{GPT-3.5 may introduce spurious supporting information}

\emph{G} sometimes introduces improbable but possible details.
For instance, \emph{G} overemphasised the significance of a patient's anxiety disorder regarding an episode of syncope and a subsequent facial fracture, which the clinicians consider unlikely.

%% Failed to describe diagnoses as interconnected events
\subsubsection{GPT-3.5 failed to present diagnoses as interconnected events}

\emph{G} does not effectively present diagnoses as interconnected, resulting in fragmented notes that lack coherence.
The clinicians described \emph{G}-generated clinical notes as collections of unrelated facts.
For example, \emph{G} presented complications of Type 1 diabetes mellitus
% (\ie ``E10.43'', ``E10.621'', ``E10.628'', ``E10.21'', ``E10.319'')
separately without illustrating their relation.
% Within the same note \emph{G} failed to recognise the connection between vascular complications and diabetes, presenting them as independent facts.
% For instance, in one of the synthetic clinical notes, ChatGPT failed to take into consideration the patient's family history which is related to their cancer diagnosis.
The lack of coherence between diagnoses may impede the plausibility of the clinical note and undermine the overall acceptability and usefulness of synthetic notes.

%% Failed to prioritise and emphasise more important diagnoses
\subsubsection{GPT-3.5 failed to prioritise and emphasise critical diagnoses}

\emph{G} struggles to prioritise diagnoses based on clinical significance, which undermines the authenticity of the portrayed scenario.
For example, \emph{G} often places critical conditions on the same level as minor issues, such as impacted ear wax, cataracts, and conjunctival haemorrhage.
Hence, we concluded that \emph{G} struggles to effectively convey the relative clinical significance of certain diagnoses.

% \section{\textbf{DISCUSSION}}
%\label{sec:dis}
%\subsection{Ethics}

%For clinical notes, we adopt an ethically approved setting, using Azure AI API with tuning off abuse monitoring and content filtering, to allow the protection of privacy as suggested by the MIMIC dataset\footnote{\url{https://physionet.org/news/post/415}}. % can put earlier in the experimental setting part
\section{\textbf{DISCUSSION, CONCLUSION AND FUTURE STUDIES}}
\label{sec:con}
In this work, we have investigated the capability of GPT-3.5's potential in augmenting ICD-10 coding for local neural models in low-resource scenarios. While overall performance dipped with synthetic data augmentation, filtered codeset evaluation showed improvements, especially in advanced models like LAAT and Multi-Res CNN. Error analysis indicated augmented models made fewer out-of-family predictions, with some shift to within-family errors (closer to the correct answer). Augmentation showed promise in improving the prediction of generated codes and their siblings. Zero-shot labels did not consistently benefit from the augmentation, emphasising the need for real data in augmentation success. However, a zero-shot code learned from the synthetic data was predicted correctly. The potential of LLM-generated discharge summaries should further be explored with different (\eg local or specialised) LLMs, prompt engineering, and further supplementing the prompt with external knowledge (\eg from ontologies).

In guided synthetic settings with ICD-10 descriptions, GPT-3.5 showed partial code identification ability displaying lesser over-/under-prediction tendencies than previously reported local models. It, however, struggled in the realistic scenario without in-prompt aid, performing below locally-trained models. Hence, the explored setup of producing a synthetic document based solely on the associated ICD codes is unsuitable for deployment in a clinical setting.

Clinician-evaluated synthetic discharge summaries showed correctness in individual codes, yet lacked naturalness and coherence compared to real data, resulting in lower informativeness, authenticity, and acceptability scores. Synthetic summaries failed to represent holistic patient narratives or prioritise critical diagnoses.

%For potential solutions
One potential solution to generating synthetic discharge summaries involves restructuring the prompt to order diagnoses chronologically, providing their corresponding timestamps.
This could guide LLMs in creating synthetic notes mirroring the chronological progression of a patient's medical journey, enhancing coherence and prioritisation.

%% In context learning
Another promising solution is to retrieve real clinical notes as in-context learning examples to help guide the generation process~\cite{lewis2020retrieval} to aid LLMs in generating more realistic and coherent content. 
As this study focuses on evaluating LLMs' existing capability, we opted to evaluate it in a zero-shot framework.
Future work may explore this idea's potential for generating more realistic-looking clinical notes.

\mf{
\section{Limitations}
\label{sec:lim}
In this study, while the annotation experts are involved as co-authors, we ensured that they were independent from the development of the algorithms that involved the synthetic data. While the evaluation utilised few clinical experts (n=4), they provided sufficient expertise in evaluating the notes. The study was blinded with respect to the real/synthetic status of documents, but according to the experts the synthetic data differed from real enough to be distinguishable.}

\section{Data Availability}

The synthetic discharge summary data generated as part of this study will be shared on reasonable request to the corresponding author upon presenting a certificate of completion of the CITI Data or Specimens Only Research course from the Collaborative Institutional Training Initiative program (\url{https://physionet.org/about/citi-course/}). The data has been accepted for publication and will be made available via PhysioNet\footnote{\url{https://doi.org/10.13026/bnc2-1a81}}.

\section{Funding}

This work is supported by the United Kingdom Research and Innovation (grant EP/S02431X/1), UKRI Centre for Doctoral Training in Biomedical AI at the University of Edinburgh, School of Informatics. HD is supported by the Engineering and Physical Sciences Research Council
(EPSRC, grant EP/V050869/1),  Concur: Knowledge Base Construction and Curation. RSP is a fellow on the Multimorbidity Doctoral Training Programme for Health Professionals, which is supported by the Wellcome Trust [223499/Z/21/Z]. BA is supported by the Advanced Care Research Centre at the University of Edinburgh.

\section{Competing Interests}
We have identified no competing interests.

\section{Author Contributions}

MF proposed exploring data augmentation for ICD coding with GPT-3.5 and developed the main ideas of the publication with APG, HD, AB, and BA. MF, APG, HD, AB, and BA contributed to the experimental design. MF designed and performed the document synthesis and the evaluation of GPT's ICD-10 coding performance on GPT-generated data. APG evaluated GPT's ICD-10 coding performance on MIMIC-IV data. MF, APG, HD, and LD designed the evaluation of discharge summaries by clinical staff. LD, SB, MH, RSP performed the clinical expert evaluation of the data, and MF and APG analysed its results. MF and APG wrote the manuscript. All authors participated in editing the manuscript. BA and AB supervised the project and provided feedback and input on the experiments and analyses.

\bibliographystyle{customunsrtnat}
\bibliography{Bibliography}
% \printbibliography



\section*{Supplementary material}

# Supplementary Materials

## 1 List of Codes Targeted in Generation

E10.3299, E10.3519, E10.3531, E10.3599, E10.52, G43.009, G43.501, G43.919,
G43.A0, G43.B0, G43.D0, H35.3110, H35.373, H81.21, H81.399, H81.8X3, H81.91,
H81.92, S00.11XA, S00.531A, S00.532A, S00.81XD, S00.93XA, S02.0XXA, S02.113A,
S02.119A, S02.31XA, S02.32XA, S02.3XXA, S02.401A, S02.402A, S02.40CA,
S02.40DA, S02.40EA, S02.40FA, S02.411A, S02.412A, S02.413A, S02.5XXA,
S02.601A, S02.609D, S02.611A, S02.61XA, S02.621A, S02.622A, S02.63XA, S02.652A,
S02.66XA, S02.81XA, S02.82XA, S02.8XXA, S06.0X9A, S06.1X0D, S06.2X0A,
S06.2X6A, S06.2X7A, S06.300A, S06.339A, S06.359A, S06.5X0D, S06.5X1A,
S06.5X7A, S06.5X9D, S06.6X1A, S06.6X6A, S06.890A, S06.9X0S, S06.9X3A,
S06.9X9A, T82.03XA, T82.09XA, T82.190A, T82.223A, T82.310A, T82.330A,
T82.338A, T82.398D, T82.49XD, T82.594A, T82.6XXA, T82.856A, T82.857A,
T82.867A, T82.868S, T84.020A, T84.023A, T84.032A, T84.033A, T84.052A,
T84.226A, T84.296A, T84.51XA, T84.53XD, T84.59XA, T84.620D, T84.623A,
T84.63XA, T84.7XXA, T84.89XA, T85.02XS, T85.43XA, T85.518A, T85.520A,
T85.528A, T85.598A, T85.611A, T85.691A, T85.694A, T85.698A, T85.71XA,
T85.72XA, T85.848A, T85.898A, T85.9XXA

# 2 Prompts for GPT-3.5

## 2.1 Generic prompt

Write a discharge summary with a detailed hospital course for a patient with these conditions:
(list of conditions)

The patient underwent these procedures:
(list of procedures)

The Discharge Summary should have a word limit of 4000 words. The names of people and locations within the discharge summary should be de-identified. Do not state ICD-10 codes in the main body of the text. For any condition involving a numeric range, state explicitly a number within that range (e.g., for a patient with blood glucose between 7.0 and 11.0 mmol/l generate "patient's glucose level was 8.6 mmol/l"). For conditions with the keyword "other" specify the condition that falls into the set of "other" - e.g., for "Other specified congenital malformations of skin" generate "Aplasia cutis congenita". For conditions with the keyword "unspecified", do not include the word "unspecified" within the main body of text - e.g., for the code H35.00 (Unspecified background retinopathy) generate the statement "the patient suffers from background retinopathy". At the end of the discharge summary add a paragraph with the header "DISCHARGE DIAGNOSES AND PROCEDURES" and assign ICD-10 codes to the discharge diagnoses and procedures, for each concept stating its ICD-10 code and its description (e.g., Essential (Primary) Hypertension [I10]).

## 2.2 Example prompt

Write a discharge summary with a detailed hospital course for a patient with these conditions:

1. Type 1 diabetes mellitus with proliferative diabetic retinopathy with traction retinal detachment not involving the macula, right eye
2. Non-pressure chronic ulcer of left heel and midfoot with unspecified severity
3. Type 1 diabetes mellitus with diabetic polyneuropathy
4. Type 1 diabetes mellitus with foot ulcer
5. Type 1 diabetes mellitus with proliferative diabetic retinopathy without macular edema, bilateral
6. Long term (current) use of insulin
7. Major depressive disorder, single episode, unspecified
8. Personal history of nicotine dependence
9. Fever, unspecified

The patient underwent these procedures:

1. Fusion of Left Tarsal Joint with Internal Fixation Device, Open Approach
2. Division of Left Foot Tendon, Open Approach
3. Insertion of External Fixation Device into Left Tibia, Percutaneous Approach

The Discharge Summary should have a word limit of 4000 words. The names of people and locations within the discharge summary should be de-identified. Do not state ICD-10 codes in the main body of the text. For any condition involving a numeric range, state explicitly a number within that range (e.g., for a patient with blood glucose between 7.0 and 11.0 mmol/l generate "patient's glucose level was 8.6 mmol/l"). For conditions with the keyword "other" specify the condition that falls into the set of "other" - e.g., for "Other specified congenital malformations of skin" generate "Aplasia cutis congenita". For conditions with the keyword "unspecified", do not include the word "unspecified" within the main body of text - e.g., for the code H35.00 (Unspecified background retinopathy) generate the statement "the patient suffers from background retinopathy". At the end of the discharge summary add a paragraph with the header "DISCHARGE DIAGNOSES AND PROCEDURES" and assign ICD-10 codes to the discharge diagnoses and procedures, for each concept stating its ICD-10 code and its description (e.g., Essential (Primary) Hypertension [I10]). Finish the document with a DISCHARGE STATUS section for the patient - either "DEAD" or "ALIVE"

## 3 Non-Hierarchical Metrics

For evaluating the model's test performance, we employed standard information retrieval metrics commonly used in LTMC tasks—micro- and macro-averaged Precision (Eq. 4), Recall (Eq. 5), and F1 scores (Eq. 6). These metrics rely on the true positives (Eq. 1), false positives (Eq. 2), and false negatives (Eq. 3),

derived from the prediction ($Z_d$) and gold standard ($Y_d$) sets for each document ($d$) in the evaluation set ($D$).

$$TP_d = |Z_d \cap Y_d| \tag{1}$$

$$FP_d = |Z_d - Y_d| \tag{2}$$

$$FN_d = |Y_d - Z_d| \tag{3}$$

$$P = \frac{\sum_{d=1}^{D} TP_d}{\sum_{d=1}^{D} TP_d + FP_d} \tag{4}$$

$$R = \frac{\sum_{d=1}^{D} TP_d}{\sum_{d=1}^{D} TP_d + FN_d} \tag{5}$$

$$F_1 = \frac{2 \cdot P \cdot R}{P + R} \tag{6}$$

Micro-averaging (*Eq.*7) assigns equal weight to each prediction, favouring high-population classes (*e.g.*, hypertension). Macro-averaging (*Eq.*8), in contrast, computes the performance for each unique label and averages across the label space, giving each label's average result equal weight regardless of their population. This highlights poor performance in less common classes. Our primary evaluation metrics common with the majority of previous work are micro-F1 (*Eq.*9) and macro-F1 scores (*Eq.*10)

$$M_{micro} = M\left(\sum_{l=1}^{|L|} TP_l, \sum_{l=1}^{|L|} FP_l, \sum_{l=1}^{|L|} FN_l\right) \tag{7}$$

$$M_{macro} = \frac{1}{|L|} \sum_{l=1}^{|L|} M(TP_l, FP_l, FN_l) \tag{8}$$

$$F_{1(micro)} = \frac{2 \cdot P_{micro} \cdot R_{micro}}{P_{micro} + R_{micro}} \tag{9}$$

$$F_{1(macro)} = \frac{2 \cdot P_{macro} \cdot R_{macro}}{P_{macro} + R_{macro}} \tag{10}$$

## 4 Hierarchical Metrics

### 4.1 Set-Based

Let $X$ represent a set of $N'$ ICD codes individually denoted as $x_1$, $x_2$, ... $x_{N'}$. Let $AN_j(x_i)$ denote a function that returns a set of all ancestors of the code $x_i$ up to the depth $j$ within the ontology. The augmented set $X_{(aug)}$ consists of the union of the original codes in $X$ and their ancestors *Eq.*11.

$$X_{(aug)} = X \cup \{AN_j(x_i)|1, ..., N'\} \tag{11}$$

In set-based hierarchical evaluation we calculate the Precision, Recall and F1 score with their standard formulae based on re-defined TP($Eq.12$), FP($Eq.13$), and FN($Eq.14$).

$$TP_{(set)d} = |Z_{(aug)d} \cap Y_{(aug)d}| \tag{12}$$

$$FP_{(set)d} = |Z_{(aug)d} - Y_{(aug)d}| \tag{13}$$

$$FN_{(set)d} = |Y_{(aug)d} - Z_{(aug)d}| \tag{14}$$

## 4.2 CoPHE

In CoPHE the ancestor labels are associated with the number of descendant codes present in the respective sets. This is handled by re-defining TP, FP, and FN per document ($d$) per code family ($c$) ($Eq.$ 15, 16, 17). The per-document value of TP, FP, and FN is the sum of the respective per-code-family metrics (example for TP in $Eq.$ 18). The $F1$ scores for CoPHE are then calculated based on the precision and recall calculated from the CoPHE versions of TP, FP, and FN. Preserving counts allows tracking whether the correct number of descendants were predicted, or whether there is under-/over-predictions (e.g., three different instances of heart disease predicted when only one was expected).

$$TP_{(CoPHE)c,d} = min(|Z_{(aug)c,d}|, |Y_{(aug)c,d}|) \tag{15}$$

$$FP_{(CoPHE)c,d} = max(|Z_{(aug)c,d}| - |Y_{(aug)c,d}|, 0) \tag{16}$$

$$FN_{(CoPHE)c,d} = max(|Y_{(aug)c,d}| - |Z_{(aug)c,d}|, 0) \tag{17}$$

$$TP_{(CoPHE)d} = \sum_{c=1}^{|C|} TP_{(CoPHE)c,d} \tag{18}$$



\end{document}